\documentclass{article}

\PassOptionsToPackage{table,xcdraw}{xcolor}

\usepackage{microtype}
\usepackage{subcaption}
\usepackage{booktabs} %

\usepackage{hyperref}

\usepackage[accepted]{icml2026_fogen}

\usepackage{amsmath}
\usepackage{amssymb}
\usepackage{mathtools}
\usepackage{amsthm}

\usepackage{multirow}
\usepackage{graphicx}

\usepackage[table,xcdraw]{xcolor}

\usepackage{upquote}
\usepackage{xurl}
\usepackage{multicol}

\usepackage[capitalize,noabbrev]{cleveref}

\usepackage{tcolorbox}
\tcbuselibrary{theorems}

\newtcbtheorem{prompt}{Prompt}%
{colback=blue!5,colframe=blue!35!black,fonttitle=\bfseries}{prompt}
\crefname{tcb@cnt@prompt}{Prompt}{Prompts}
\Crefname{tcb@cnt@prompt}{Prompt}{Prompts}

\DeclareCaptionType{mybox}[Box][List of Boxes]
\newtcbtheorem[no counter]{entitybox}{}%
  {colback=gray!5, colframe=black!60, fonttitle=\bfseries,
   separator sign={}}{box}
\crefname{mybox}{Box}{Boxes}
\Crefname{mybox}{Box}{Boxes}

\usepackage{enumitem}
\usepackage{pifont}

\newcommand{\yes}{{\small\color{green!50!black}\ding{51}}}
\newcommand{\no}{{\small\color{red!70!black}\ding{55}}}

\newcommand{\qa}[2]{%
\begin{tabular*}{\linewidth}{@{\extracolsep{\fill}}p{0.92\linewidth}r}
#1 & #2
\end{tabular*}%
}

\theoremstyle{plain}

\theoremstyle{definition}

\theoremstyle{remark}

\icmltitlerunning{Amortising Bayesian Experimental Design for Information Gathering in LLMs}

\begin{document}

\noindent{\small ICML 2026 Workshop on Foundations of Deep Generative Models:
Understanding Memorization, Generalization, and Reasoning}
\vskip 0.35in

\icmltitle{Amortising Bayesian Experimental Design \\ for Sequential Information Gathering in LLMs}

\icmlsetsymbol{equal}{*}

\begin{icmlauthorlist}
    \icmlauthor{Jakob Hartmann}{equal,oxford}
    \icmlauthor{James Harvey}{equal,oxford}
    \icmlauthor{Jhonathan Navott}{equal,oxford}
    \icmlauthor{Erik Y. Wang}{oxford}
    \icmlauthor{Luckeciano C. Melo}{oxford}
    \icmlauthor{Flaviu Cipcigan}{eit}
    \icmlauthor{Cheng Zhang}{eit}
    \icmlauthor{Alessandro Abate}{oxford}
\end{icmlauthorlist}

\icmlaffiliation{oxford}{University of Oxford, Oxford, United Kingdom}
\icmlaffiliation{eit}{Ellison Institute of Technology, Oxford, United Kingdom}

\icmlcorrespondingauthor{Jakob Hartmann}{jakob.hartmann@dtc.ox.ac.uk}

  \icmlkeywords{Large Language Models, Bayesian Experimental Design, Sequential Information Gathering, Expected Information Gain, Multi-Turn Reasoning, Group Relative Policy Optimisation}

\vskip 0.3in

\printAffiliationsAndNotice{\icmlEqualContribution}

\begin{abstract}
  Large language models (LLMs) exhibit strong reasoning and world-knowledge capabilities, yet often struggle to gather information effectively across the multi-turn interactions required in sequential decision-making settings. We introduce \textbf{A}mortised \textbf{S}equential \textbf{I}nformation \textbf{G}athering (ASIG), a fine-tuning approach that amortises Bayesian Experimental Design (BED) into LLM policies via a multi-turn extension of Group Relative Policy Optimisation with an Expected Information Gain reward. Evaluated on the 20 Questions task, ASIG more than doubles the success rate of the 7B base model and reduces inference cost by over $25\times$ relative to BED-LLM, a competitive inference-time baseline. Applied to MediQ, a medical diagnosis benchmark unseen during training, ASIG improves information-seeking performance at the 7B scale, suggesting that the learned strategies can transfer out of distribution. Our findings show that amortising BED into LLM policies provides an effective and computationally efficient approach to sequential information gathering.
\end{abstract}

\section{Introduction}
Effective information gathering in sequential settings is fundamental to applications ranging from healthcare to designing experiments in scientific domains. LLMs are increasingly being deployed in such domains, as they are high capacity models encoding extensive world knowledge \cite{petroniLanguageModelsKnowledge2019}, and have shown promise in reasoning and decision-making tasks on diverse action spaces expressed in natural language \cite{yaoReActSynergizingReasoning2023}. However, despite these capabilities, LLMs have been shown to struggle with multi-turn reasoning and planning in various settings. One example is the ``lost in conversation" phenomenon reported by \citet{labanLLMsGetLost2026}, which demonstrates that LLMs often struggle to (1) recognize when additional information is needed to complete a task and (2) effectively acquire that information during multi-turn conversations. These limitations are particularly problematic for experimental design and sequential decision making, where maintaining and updating coherent beliefs over extended interactions is key.

A growing body of work has explored methods for improving the decision-making and information-gathering capabilities of LLMs in sequential settings. Broadly, there are two main classes of approaches: \textit{weight-space optimisation} methods \cite{mazzaccaraLearningAskInformative2024, tajwarTrainingGenerallyCurious2025, chiCLARINETAugmentingLanguage2024, yunSherlockYourQueries2025, auzinaIntrinsicCreditAssignment2026}, which fine-tune the model parameters, and \textit{inference-time optimisation} methods \cite{choudhuryBEDLLMIntelligentInformation2025, kobalczykActiveTaskDisambiguation2025, grandShootFirstAsk2025}, which add post-processing procedures during inference. While demonstrating strong improvement in the information-gathering performance of LLMs in sequential settings, inference-time optimisation methods suffer from high computational cost during inference and must be tailored to a specific task or environment. On the other hand, weight-space optimisation methods yield fast inference time behaviour, but it remains to be seen whether these approaches improve general information gathering capabilities or just fit to the task at hand. 

In this work, we introduce Amortised Sequential Information Gathering (ASIG), a weight-space optimisation framework that amortises Bayesian Experimental Design (BED) \cite{chalonerBayesianExperimentalDesign1995} into the weights of an LLM, enabling sequential information gathering without additional inference-time optimisation. We treat the LLM as a BED agent in a general sequential setting, where it iteratively selects experiments to maximally reduce uncertainty over a latent target. We then fine-tune the model using Expected Information Gain (EIG) as the reward signal within a multi-turn variant of Group Relative Policy Optimisation (GRPO) \cite{shaoDeepSeekMathPushingLimits2024}. The reward combines EIG, which incentivises experiments that maximally reduce uncertainty over candidate targets, with an outcome-based signal that directly reinforces task success, trained via a learning curriculum across multi-turn rollouts. As a controlled instantiation of this framework, we train on the 20 Questions game, where binary outcomes make EIG tractable via a Rao-Blackwellised estimator and the capability being trained (asking maximally informative questions) is inherently domain-agnostic. We evaluate on held-out 20 Questions categories and assess generalisation to MediQ \cite{liMediQQuestionAskingLLMs2024}, exploring whether amortised BED behaviour transfers to real-world sequential information-gathering settings.

\begin{figure*}[t]
    \centering
    \includegraphics[width=\textwidth]{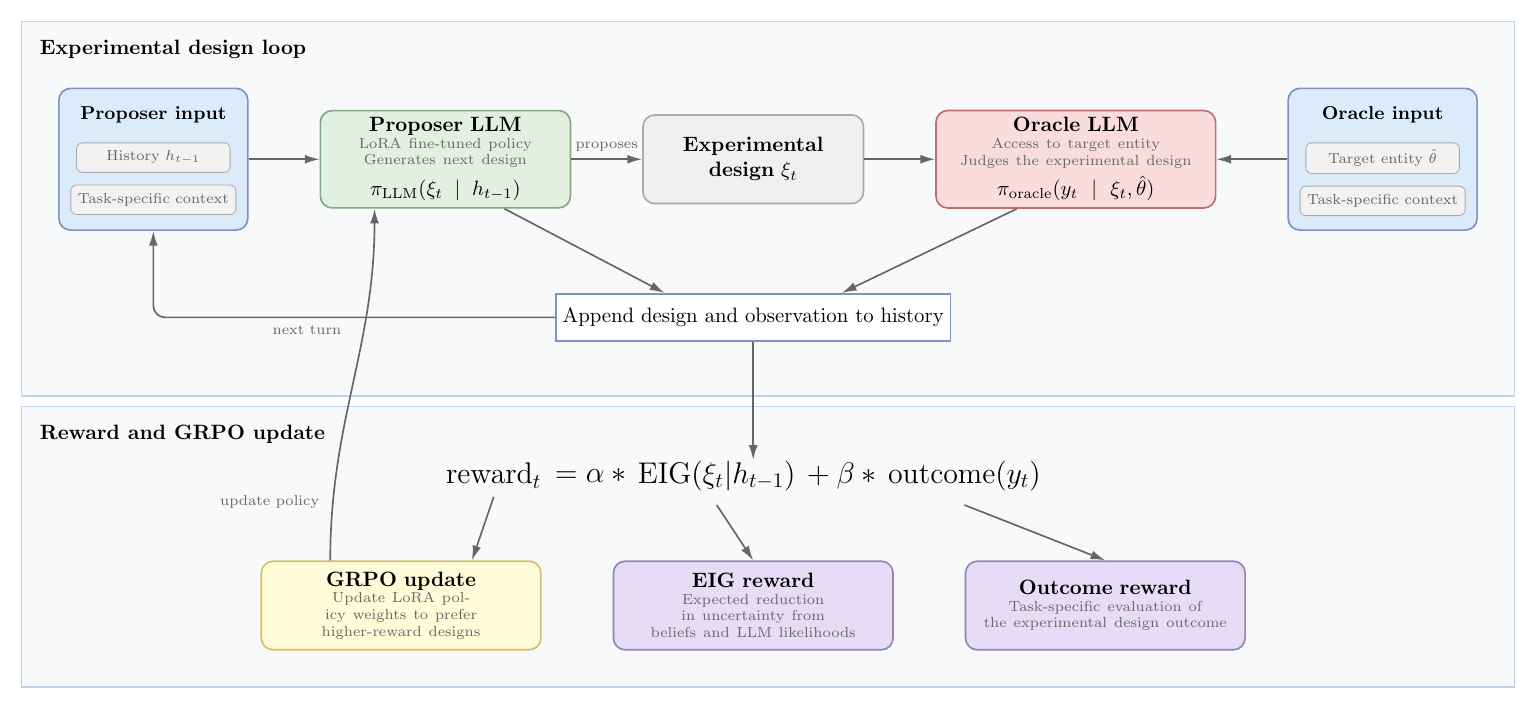}
    \caption{\textbf{High-level overview of our experimental design loop.} At each turn, the Proposer LLM uses the task-specific context and interaction history to generate a new experimental design. The Oracle LLM, which has access to the hidden target entity, evaluates this design and returns an observation. Both the design and observation are appended to the history for the next turn. Finally, we compute a reward combining the Expected Information Gain (EIG) and a task-specific outcome score, using GRPO to update the Proposer LLM's policy weights.}
    \label{fig:method}
\end{figure*}

We summarise our contributions as follows:
\begin{enumerate}
    \item We introduce ASIG, a novel fine-tuning method to amortise BED information gathering into LLMs.
    \item We extend GRPO to support multiple reward streams with distinct temporal horizons, enabling joint optimisation of turn-level EIG and multi-horizon task success. 
    \item We demonstrate that ASIG substantially improves 20 Questions performance over the base model, more than doubling the success rate in the 7B model class, while outperforming a competitive inference-time baseline in two out of four settings at over 25× lower inference cost.
    \item We show that the information-seeking behaviour learned by ASIG transfers to MediQ, a medical diagnosis benchmark, at the 7B scale without any in-domain fine-tuning.
\end{enumerate}

\section{Related Work}
\subsection{Inference-Time Optimisation Methods}
\label{subsec:inference_time_optimisation_methods}
Inference-time optimisation methods improve sequential decision making and question asking in LLMs by optimising information gathering directly at test time, without modifying model weights. In this setting, LLMs are treated as decision making agents in a sequential problem with a hidden target, where the next action is selected by explicitly maximizing EIG (or an EIG proxy) given the interaction history. BED-LLM \cite{choudhuryBEDLLMIntelligentInformation2025} adopts this approach on the 20 Questions task. They use the LLM to propose candidate questions, define a uniform prior over the set of possible targets, and use the questions and possible targets to estimate EIG, eventually selecting the question with the highest score at each turn. Related methods adopt similar formulations, using sampling based EIG estimation for task disambiguation \cite{kobalczykActiveTaskDisambiguation2025}, or maintaining explicit Bayesian belief states with Monte Carlo estimation of the EIG \cite{grandShootFirstAsk2025}.

These approaches consistently outperform naive prompting, offering principled uncertainty handling and interpretability through explicit belief updates, while remaining applicable to off-the-shelf LLMs. However, they incur substantial inference time cost due to sampling and belief maintenance, often rely on structured hypothesis spaces or simulators (e.g. 20 Questions-type tasks), and are typically myopic and task-specific which limits their actual usability.

\subsection{Weight-Space Optimisation Methods} 
Weight-space methods for improving LLM information-gathering divide broadly into retriever-coupled and self-contained approaches. \citet{chiCLARINETAugmentingLanguage2024} and \citet{yunSherlockYourQueries2025} both couple the LLM to an external retriever network that carries the belief state, deriving EIG or rank-based rewards from the retriever's candidate distribution. However, neither evaluates transfer beyond their training tasks, and both are restricted to settings where retrieval infrastructure exists.

Self-contained approaches remove this dependency. \citet{mazzaccaraLearningAskInformative2024} demonstrate that EIG is an effective training signal via Direct Preference Optimisation (DPO) \cite{rafailovDirectPreferenceOptimization2023} on a static dataset of EIG-ranked question pairs, but the fixed preference dataset cannot adapt to the model's evolving question distribution, and generalisation is evaluated only within 20 Questions categories. \citet{tajwarTrainingGenerallyCurious2025} takes a complementary approach, using Relative Preference Optimisation (RPO) \cite{pangIterativeReasoningPreference2024} on purely outcome-based rewards with no information-theoretic component. They demonstrate zero-shot transfer across a family of simple decision-making tasks, though the authors themselves note that online RL would likely yield stronger results than their offline preference optimisation.

$\Delta$Belief-RL \cite{auzinaIntrinsicCreditAssignment2026} is the closest antecedent to our work: an online RL method that uses the change in the model's assigned probability to the correct target as a dense per-turn reward, without any external retriever, and demonstrates OOD generalisation to customer service and personalisation settings. Our approach shares the same self-contained online RL paradigm, but uses EIG instead of $\Delta\text{Belief(target)}$ as the reward signal. This distinction matters because EIG captures both the uncertainty of the marginal answer distribution and the sharpness of per-belief likelihoods, rather than tracking a single scalar probability whose change may not reflect genuine information gain.

\section{Preliminary}
\subsection{Bayesian Experimental Design}
BED provides a principled framework for experimental design grounded in Bayesian decision theory, in which experiments are selected based on their expected utility under an explicit probabilistic model of the world \cite{chalonerBayesianExperimentalDesign1995}. In the BED formulation, the uncertainty about a quantity of interest $\theta$ is represented by a prior distribution $p(\theta)$, while a likelihood model $p(y |\theta, \xi)$ specifies how experimental outcomes $y$ are generated under a proposed design $\xi$. Together, these components define a joint probabilistic model over latent variables and observations, enabling candidate experiments to be evaluated prior to data collection by computing their expected utility. This explicit generative modelling distinguishes BED from heuristic or purely discriminative approaches to experimental design, and allows prior knowledge, measurement noise, and experimental constraints to be incorporated in a coherent Bayesian manner \cite{rainforthModernBayesianExperimental2024}.

\begin{figure}[t]
    \centering
    \includegraphics[width=\textwidth]{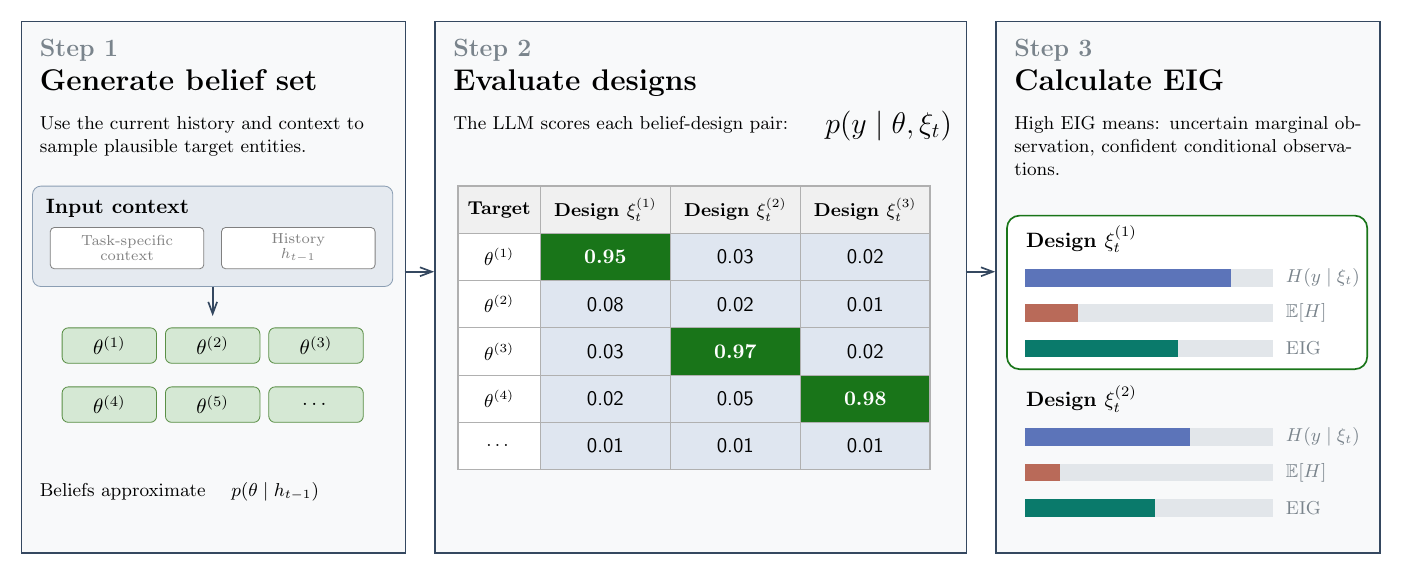}
    \caption{\textbf{Overview of our step-by-step EIG calculation, following the approach of BED-LLM \cite{choudhuryBEDLLMIntelligentInformation2025}.} 1) Plausible target entities ($\theta$) are sampled from the current belief state. 2) For each design ($\xi_t$), observation likelihoods $p(y \mid \theta, \xi_t)$ are computed across all sampled targets. 3) These likelihoods are used to calculate the predictive entropy $H(y_t \mid \xi_t)$ and expected conditional entropy $\mathbb{E}[H(y_t|\theta,\xi_t)]$. High-EIG designs are those that produce uncertain marginal observations (splitting the belief space) while minimizing conditional entropy.}
    \label{fig:eig}
\end{figure}

\paragraph{Expected Information Gain.}
First proposed by \citet{lindleyMeasureInformationProvided1956}, EIG is a widely used optimality metric within BED. Given a latent variable $\theta$, which we seek to learn about through an experiment $\xi$, we define the \textit{Information Gain} (IG) on $\theta$ from an experimental outcome $y$ by the reduction in Shannon entropy \cite{shannonMathematicalTheoryCommunication1948} from the prior to the posterior:

\begin{equation}
    \operatorname{IG}_{\theta}(\xi) = H[p(\theta)] - H[p(\theta|y,\xi)].
\end{equation}
EIG is then calculated as the expectation of this quantity over the marginal predictive distribution, $p(y|\xi)$:
\begin{align}
    \operatorname{EIG}_{\theta}(\xi) &= \mathbb{E}_{p(y|\xi)}[\operatorname{IG}_{\theta}(\xi,y)] \label{eq:2} \\
    &= \mathbb{E}_{p(\theta)p(y|\theta,\xi)}[\log{p(\theta|y,\xi)}-\log{p(\theta)}]\notag \\
    &= \mathbb{E}_{p(\theta)p(y|\theta, \xi)}[\log{p(y|\theta,\xi)}-\log{p(y|\xi)}].\notag
\end{align}

Finding the optimal experimental design thus amounts to maximising $\operatorname{EIG}_{\theta}(\xi)$ with respect to $\xi$. However, as both $p(y|\xi)$ and $p(\theta|y, \xi)$ are intractable, EIG cannot be directly estimated from traditional Monte Carlo (MC) estimators \cite{rainforthModernBayesianExperimental2024, fosterUnifiedStochasticGradient2020}. However, \citet{rainforthNestingMonteCarlo2018} point out that when the outcome space $\mathcal Y$ is finite, EIG can be estimated using a Rao-Blackwellised estimator:

\begin{equation} \label{eq: rao-blackwell}
    \hat{\mu}_N=\sum_{y\in\mathcal{Y}}\bigg[\frac{1}{N}\sum_{n=1}^N p(y|\theta_n,\xi)\log{p(y|\theta_n,\xi)} -\hat{p}(y|\xi)\log{\hat{p}(y|\xi)} \bigg],
\end{equation}

where $\hat{p}(y|\xi)=\frac{1}{N}\sum_{n=1}^Np(y|\theta_n,\xi)$. Since this is essentially a conventional Monte Carlo estimator with no nesting, it has the advantage of sharing the same mean squared error convergence rate ($\mathcal{O}(1/N)$) as in standard Monte Carlo.

\paragraph{Sequential BED.}
BED can be extended to sequential design settings by iteratively incorporating information gathered by prior experiments to select the next one \cite{rainforthModernBayesianExperimental2024}. Formally, given a history $h_{t-1}=\{(\xi_k,y_k)\}^{t-1}_{k=1}$, the EIG of design $\xi_t$ is given by\footnote{This assumes that $y_t$ is independent of $h_{t-1}$ given $\theta$ and $\xi_t$, which is the case for our 20 Questions training environment.}

\begin{equation}
\operatorname{EIG}_{\theta}(\xi_t|h_{t-1}) = \mathbb{E}_{p(\theta|h_{t-1})p(y_t|\theta,\xi_t)}\bigg[\log\frac{p(y_t|\theta,\xi_t)}{p(y_t|\xi_t)} \bigg],\notag
\end{equation}

where $h_0=\varnothing$. This corresponds to updating the prior with each observation to select the next design.

\subsection{Group Relative Policy Optimisation}
GRPO \cite{shaoDeepSeekMathPushingLimits2024} is a variant of Proximal Policy Optimization (PPO) \cite{schulmanProximalPolicyOptimization2017} that eliminates the need for an expensive critic model to approximate the value function and instead calculates the advantage using the relative differences between individual rollout rewards and the group mean. At each training step, GRPO samples a prompt from the prompt dataset, performs $G$ rollouts, and then scores them with a given reward model. These advantages are then used to update the policy $\pi_{\text{LLM}}$ using a clipped surrogate objective.

\section{ASIG: Amortised Sequential Information Gathering}
\label{sec:method}

\begin{figure}[t]
    \centering
    \includegraphics[width=\textwidth]{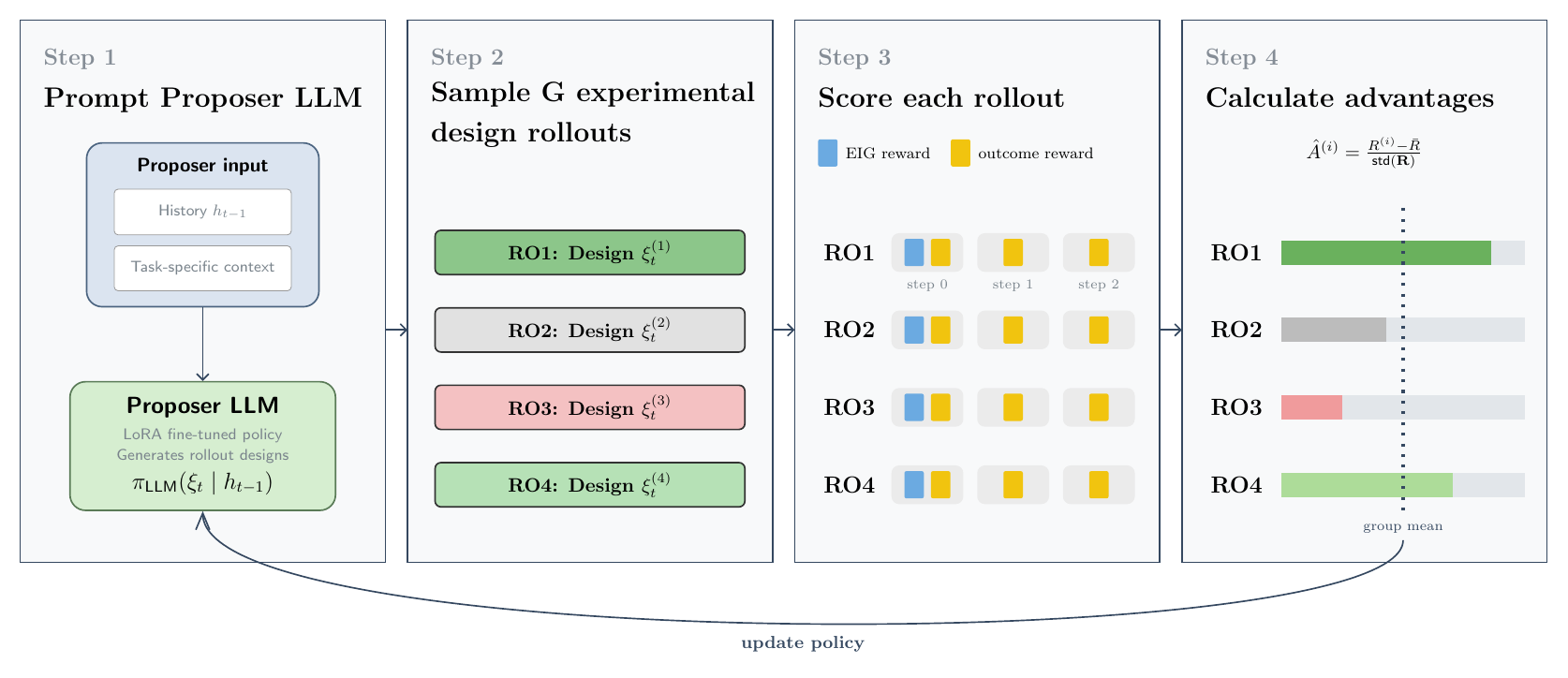}
    \caption{\textbf{Overview of our multi-turn GRPO method.} At each turn, the Proposer LLM generates $G$ rollouts based on the given context and interaction history. Each rollout consists of multiple steps of experimental design and observation. We calculate the EIG reward for the first step, and the outcome-based reward across all $O$ steps. The combined reward is used to calculate the group mean, and the individual advantages are then used to update the Proposer LLM’s policy.}
    \label{fig:grpo}
\end{figure}

We consider a multi-turn interaction environment, in which the Proposer LLM agent is tasked with uncovering a latent target quantity $\hat{\theta}\in\Theta$ through a budget of $T$ sequential experimental designs $\xi\in\Xi$. The environment is initialised with a task-specific prompt which defines the search space. Over a trajectory $\tau=\{(\xi_k,y_k)\}^{k=T}_{k=1}$, the agent interacts with the environment, modelled by the frozen and independent Oracle LLM, through experimental designs $\xi$ and receives its outcomes $y\in\mathcal{Y}$. The interaction history at step $t$ is denoted by $h_{t-1}=\{(\xi_k,y_k)\}^{k=t-1}_{k=1}$, with $h_0$ being the initial task-specific prompt. In this setting, we treat the Proposer LLM agent as our sequential BED policy $\pi_{\text{LLM}}(\cdot|h_{t-1})$, from which experiments are sampled based on the accumulated context. We then use the Proposer LLM to generate beliefs and compute likelihoods $p_{\text{LLM}}(y_t|\theta, \xi_t)$ to estimate the EIG for each proposal. An outcome-based reward is computed from the corresponding experimental outcomes, and together these two components define the reward signal in our multi-turn GRPO fine-tuning setup. Figure~\ref{fig:method} provides a high-level overview of the approach, while the following subsections detail the specific implementation evaluated in this work.

\subsection{Training Environment}
In this study, we implement our training procedure using the 20 Questions game environment, which has several desirable properties in this context. First, 20 Questions forms a natural instantiation of the BED framework in which the questions posed by the LLM are themselves the experimental designs $\xi_t$, and the answers are the corresponding experimental outcomes $y_t$. Additionally, the binary experimental outcome space (i.e. “yes” or “no”) allows for an efficient estimation of the EIG through the Rao-Blackwellised estimator in \cref{eq: rao-blackwell}. Finally, \citet{zhangProbingMultiturnPlanning2024} have shown that succeeding in 20 Questions requires good planning to efficiently partition the hypothesis space before deductive reasoning can be used to produce educated guesses in later turns. This generic planning-oriented structure makes the task a useful proxy for learning capabilities that may transfer to other information gathering tasks.

The resulting environment is implemented as an interaction between the Proposer LLM, which is fine-tuned, and a separate frozen Oracle LLM. At each turn, the Proposer receives the task category together with the accumulated question-answer history and generates a question or guess. The Oracle, which has access to the hidden target entity, evaluates this proposal and returns a binary answer, either ``Yes'' or ``No'', unless the Proposer has correctly identified the target, in which case the Oracle returns ``Correct'' and the interaction terminates. For non-terminal turns, the resulting question-answer pair is appended to the history and the interaction continues until either the target is identified or the budget of $T=20$ turns is exhausted.

\subsection{Reward}
\label{subsec:reward}
Effective information gathering requires the ability to both design maximally informative experiments and identify when enough information has been acquired to identify the hidden entity and act optimally. We argue that these are two distinct skills, which motivates two separate components for the reward signal. For each experimental design (e.g. a question in 20 Questions) generated in the GRPO rollout, we calculate two reward components: EIG and outcome reward. These two signals are complementary. EIG provides dense, turn-level feedback on question quality but carries no signal on whether information gathering ultimately succeeds, while the outcome-based reward drives goal-directed behaviour but suffers from sparse credit assignment in long-horizon settings. For 20 Questions specifically, the outcome reward encourages the model to guess the hidden entity using as few questions as possible, while EIG encourages asking maximally informative questions.

The outcome reward, denoted $r_{\mathrm{out}}(y_t)$, consists of a step-wise penalty for all questions that lead to a non-terminal state and a large positive reward if the questioner has correctly guessed the hidden entity.

To calculate the EIG for a given question $\xi_t$, we use the approach introduced by \citet{choudhuryBEDLLMIntelligentInformation2025} and visualized in Figure~\ref{fig:eig}. At each turn, we sample beliefs $\theta\sim p_{\text{LLM}}(\cdot|h_{t-1})$ from the Proposer LLM to create a belief set based on the current conversation history. This set is then filtered to ensure that only coherent beliefs, which are consistent with the previous questions and answers, are retained. If the resulting belief set is too small, we perform up to three rounds of belief regeneration to ensure a large enough set. We denote the resulting belief set at turn $t$ formed from this process as $\Theta_t$. Afterwards, we uniformly sample $N$ beliefs $\theta_n\in\Theta_t$ and query the Proposer LLM to obtain the likelihoods for each sampled belief, $p_{\text{LLM}}(y_t|\theta_n, \xi_t)$, which are used to estimate the question's EIG via the Rao-Blackwellised estimator in \cref{eq: rao-blackwell}. We scale the EIG reward by $1 / \ln(2)$ to obtain a normalized scalar, $r_{\mathrm{EIG}}(\xi_t,\Theta_t)$, within the $[0,1]$ range.

\subsection{Multi-Turn GRPO}
\label{subsec:multi-turn-grpo}
We extend the default GRPO formulation \cite{shaoDeepSeekMathPushingLimits2024} to the multi-turn setting. At each training step, a starting state consisting of the task category and the interaction history up to that turn, $h_{t-1}$, is sampled from the prompt dataset outlined in \cref{subsection: prompt dataset}. From this state, the Proposer LLM generates $G$ independent policy rollouts by sampling multiple consecutive question-answer pairs within each rollout. A high-level overview is shown in Figure~\ref{fig:grpo}. For each rollout, we compute the scalar reward described in Section~\ref{subsec:reward} using 

\begin{equation}
  R_t =
    \frac{\alpha}{\mathcal{Z}_E}
    \sum_{k=0}^{E-1} \gamma_{\mathrm{EIG}}^{k}\,
      r_{\mathrm{EIG}}(\xi_{t+k},\,\Theta_{t+k})
    + \frac{\beta}{\mathcal{Z}_O}
    \sum_{k=0}^{O-1} \gamma_{\mathrm{out}}^{k}\,
      r_{\mathrm{out}}(y_{t+k}),
\end{equation}

where $\alpha, \beta$ are weighting constants, $\gamma_{\mathrm{EIG}}, \gamma_{\mathrm{out}} \in (0,1]$ are discount factors, and $E, O$ correspond to the rollout horizon for the EIG and outcome-based reward, respectively. The normalization constants $\mathcal{Z}_E$ and $\mathcal{Z}_O$ ensure that different rollout horizons for EIG and outcome-based rewards do not lead to varying reward magnitudes. For a horizon $H$ and discount factor $\gamma$, we define

\begin{equation}
\mathcal{Z}(H,\, \gamma) = \sum_{k=0}^{H-1} \gamma^{k}
  = \begin{cases}
      H & \text{if } \gamma = 1 \\[4pt]
      \dfrac{1 - \gamma^{H}}{1 - \gamma} & \text{otherwise}.
    \end{cases}
\end{equation}

In our experiments, we use $E = 1$ and $O = 3$, i.e. the EIG reward is calculated only with respect to the first question in each rollout, whereas the outcome reward also considers the two subsequent questions. This favours questions that maximize EIG at the current turn, while increasing the probability of a correct guess within three turns. After obtaining the rewards for each question, we calculate the average group return and individual rollout advantages, which are used to update the LoRA \cite{huLoRALowRankAdaptation2021} adapters of the questioner model.

\subsection{Learning Curriculum} In preliminary experiments, we observed that the Proposer LLM initially rarely makes specific guesses. To encourage this behaviour, we employ a learning curriculum strategy in which the model is first trained without the EIG reward ($\alpha = 0$), before subsequently introducing the EIG signal. This curriculum increases the frequency of hypothesis-seeking questions, which contain specific guesses, relative to constraint-seeking questions that probe the characteristics of the hidden entity.

\subsection{GRPO Prompt Dataset} 
\label{subsection: prompt dataset}
At the start of each training episode, we randomly sample $n$ target entities from a category-entity list to generate a 20 Questions self-play dataset between the Proposer and Oracle. Each state in this dataset, consisting of the entity’s category and the Q/A history up to question $t-1$, then serves as input for GRPO rollouts starting at question $t$.

\begin{figure}[t]
    \centering
    \includegraphics[width=\textwidth]{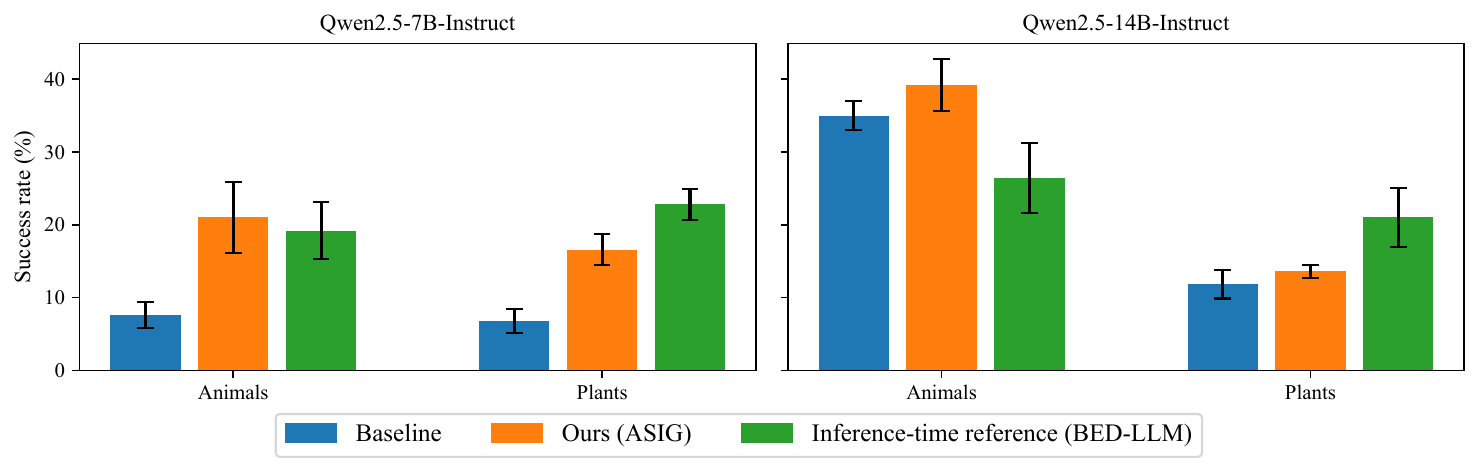}
    \caption{\textbf{Mean success rate (±1 std) over 5 episodes of Qwen2.5-7B-Instruct and Qwen2.5-14B-Instruct on the 20 Questions task across the animals and plants categories.} On the 7B model, ASIG more than doubles the success rate of the baseline and outperforms BED-LLM in the animals category. On the 14B model, ASIG improves over the baseline by 4.2 and 1.8 percentage points on animals and plants, respectively. BED-LLM performs better than ASIG on the 14B plants category but falls substantially below both ASIG and the baseline on animals, suggesting limited robustness across settings despite using an order of magnitude more inference-time compute.}
    \label{fig:20q_results}
\end{figure}

\section{Experiments}
\subsection{Setup}
To evaluate our method, we train on the 20 Questions task and assess both in-distribution and out-of-distribution performance. We use Qwen2.5-7B-Instruct and Qwen2.5-14B-Instruct as Proposer LLMs and Qwen2.5-72B-Instruct as the Oracle LLM \cite{qwenQwen25TechnicalReport2025}. For training, we construct a list of 1,000 target entities spanning 20 categories. This list was generated with Claude Opus 4.7 \cite{anthropicClaude47Opus2026} using \cref{prompt:training_dataset_generation} and manually reviewed to ensure quality. We train a LoRA adapter on this dataset following the approach in \cref{sec:method}, with hyperparameters listed in \cref{tab:hyperparameters}.

We evaluate in-distribution performance on two held-out categories and out-of-distribution performance on MediQ \cite{liMediQQuestionAskingLLMs2024}, a medical information-seeking benchmark. We further verify that improved information-seeking capabilities do not come at the cost of degraded general language abilities, as measured on three standard LLM benchmarks.

\subsection{20 Questions}
We evaluate the fine-tuned models on two held-out categories, animals and plants, each comprising 100 entities not seen during training. The animal entities are drawn from \citet{choudhuryBEDLLMIntelligentInformation2025} and the plant entities are a subset of those used by \citet{auzinaIntrinsicCreditAssignment2026}. The full lists are provided in \cref{box:animals_dataset,box:plants_dataset}. We compare against the instruction-tuned base model and our reimplementation of BED-LLM \citep{choudhuryBEDLLMIntelligentInformation2025,BEDLLMreproductionModelpyMain}.

\paragraph{Success rate.} \cref{fig:20q_results} shows the success rate across model sizes and evaluation categories, measured as the fraction of games in which the questioner identifies the hidden entity within the 20 question budget. On the 7B model, ASIG more than doubles the success rate over the baseline and surpasses BED-LLM on the animals category. On the 14B model, ASIG improves over the baseline by 4.2 and 1.8 percentage points on animals and plants, respectively. BED-LLM performs better on the 14B plants category but degrades on the animals category, falling below both ASIG and the baseline, suggesting that it does not generalise robustly across settings despite requiring an order of magnitude more inference-time compute. \cref{fig:20q_cumulative_accuracy} in the Appendix further shows that ASIG identifies the target entity earlier than the baseline, requiring fewer questions on average. Qualitative examples comparing baseline and ASIG on the same target entity are shown in \cref{box:game_fir,box:game_honey_bee} in the Appendix.

\begin{figure}[t]
    \centering
    \includegraphics[width=\columnwidth]{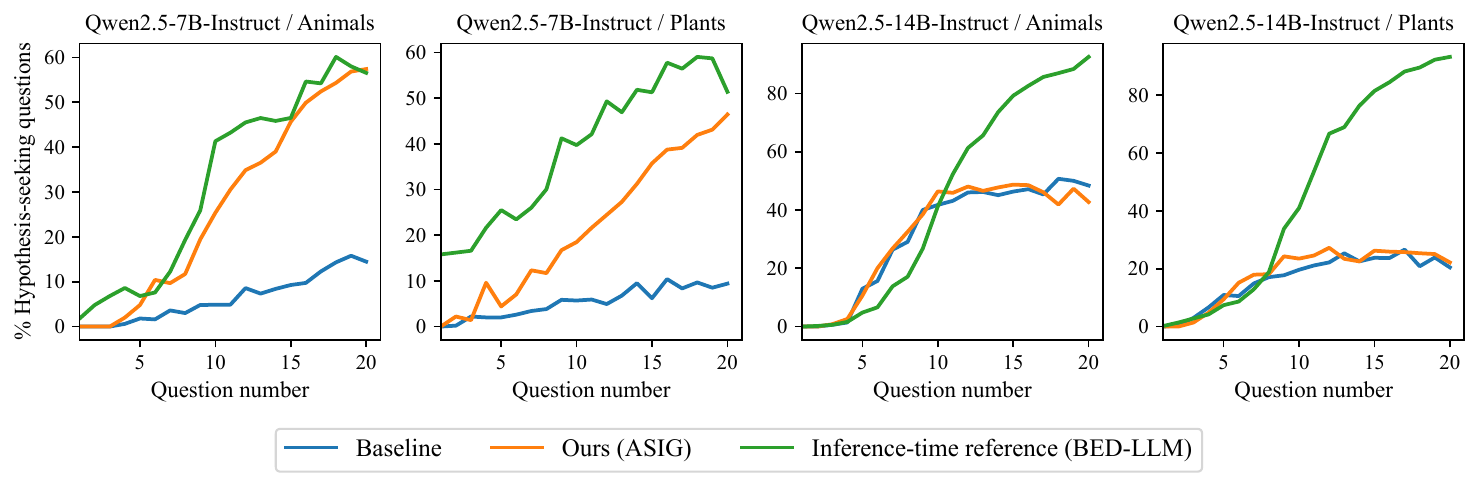}
    \caption{\textbf{Fraction of questions classified as hypothesis-seeking at each turn for Qwen2.5-7B-Instruct and Qwen2.5-14B-Instruct across the animals and plants categories.} All methods become more hypothesis-seeking as the game progresses. For the 7B model, ASIG consistently and substantially exceeds the baseline, suggesting that fine-tuning encourages earlier explicit guesses. For the 14B model, ASIG's hypothesis-seeking rate closely tracks the baseline across both categories, indicating a weaker behavioural shift. BED-LLM shows the highest rate overall, though this partly reflects its hard-coded guessing rules rather than an emergent strategy as in ASIG.}
    \label{fig:20q_hypothesis_seeking}
\end{figure}

\paragraph{Behavioural analysis.} To better understand the strategies learned by ASIG, we classify each question as either constraint-seeking (asking about the entity's characteristics to narrow down candidates) or hypothesis-seeking (making a specific guess about the target), following the taxonomy of \citet{mazzaccaraLearningAskInformative2024}. \cref{fig:20q_hypothesis_seeking} shows the fraction of hypothesis-seeking questions across turns. For the 7B model, ASIG consistently exceeds the baseline across both categories, committing to explicit guesses earlier in the game.

For the 14B model, ASIG's hypothesis-seeking rate closely tracks the baseline. This likely reflects the stronger prior capabilities of the 14B model, which requires less behavioural adaptation to benefit from fine-tuning.
BED-LLM achieves the highest overall rate, but this is partly driven by hard-coded guessing logic that forces explicit guesses once the belief set falls below three candidates, which ASIG does not rely on.

\begin{figure}[b]
    \centering
    \includegraphics[width=0.75\textwidth]{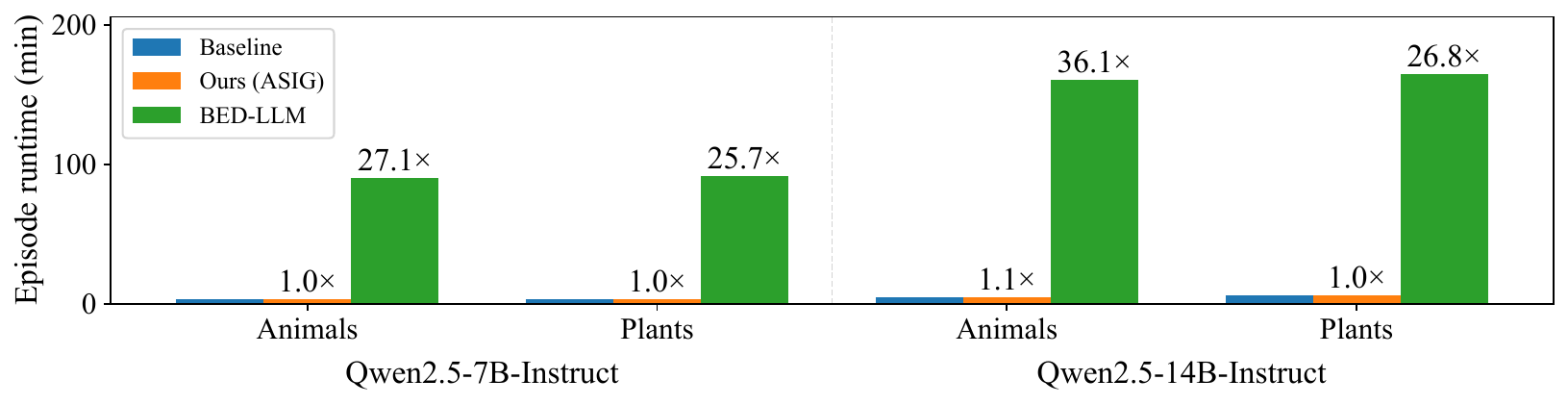}
    \caption{\textbf{Mean per-episode wall-clock runtime of the baseline, ASIG, and BED-LLM across all model and category settings.} Runtime ratios relative to the baseline are annotated above each ASIG and BED-LLM bar. ASIG matches the baseline runtime, while BED-LLM is between 25× and 36× slower, as it requires repeated belief sampling and EIG estimation at every turn rather than a single forward pass.}
    \label{fig:20q_runtime}
\end{figure}

\begin{table}[t]
    \centering
    \caption{\textbf{MediQ results for the baseline and ASIG on Qwen2.5-7B-Instruct and Qwen2.5-14B-Instruct.} For the 7B model, ASIG improves accuracy and voluntary accuracy while reducing overconfidence, at a minor cost in underconfidence. Results are more mixed for the 14B model, consistent with the weaker behavioural adaptation observed in the hypothesis-seeking analysis.}
    \label{tab:mediq_results}
    \resizebox{\textwidth}{!}{%
\begin{tabular}{@{}llllll@{}}
\toprule
Model                                 & Variant     & Accuracy (\%) $\uparrow$ & Voluntary accuracy (\%) $\uparrow$ & Overconfidence rate (\%) $\downarrow$ & Underconfidence rate (\%) $\downarrow$ \\ \midrule
\multirow{2}{*}{Qwen2.5-7B-Instruct}  & Baseline    & 46.7                     & 50.1                               & 36.1                                  & \textbf{10.6}                          \\
                                      & Ours (ASIG) & \textbf{47.6}            & \textbf{51.8}                      & \textbf{33.4}                         & 11.7                                   \\
\multirow{2}{*}{Qwen2.5-14B-Instruct} & Baseline    & 54.1                     & \textbf{64.6}                      & \textbf{20.7}                         & 16.5                                   \\
                                      & Ours (ASIG) & \textbf{54.2}            & 63.0                               & 22.2                                  & \textbf{16.4}                          \\ \bottomrule
\end{tabular}%
}

\end{table}

\paragraph{Inference-Time Efficiency.} By amortising the Bayesian experimental design computation into the model weights, ASIG incurs the same inference cost as the base LLM. Notably, this makes ASIG between 25x and 36x faster than BED-LLM (\cref{fig:20q_runtime}), which must perform explicit belief sampling and EIG maximisation at every turn. This efficiency gain makes ASIG considerably more applicable to real-world settings where latency and compute are constrained.

\subsection{Generalization to Medical Information Seeking}
To assess out-of-distribution performance, we evaluate our fine-tuned models on MediQ \citep{liMediQQuestionAskingLLMs2024}, a benchmark that simulates a clinical interaction in which a doctor elicits information from a patient through follow-up questions before answering a multiple-choice clinical question. At each turn, a doctor agent uses an abstention module to decide whether sufficient information has been gathered to answer or whether to ask another question. We use our fine-tuned Qwen2.5-7B and Qwen2.5-14B models from the 20 Questions task as the doctor and Qwen2.5-72B-Instruct as the patient, which responds based on a provided patient record. 

We compare each fine-tuned model to its base counterpart across four metrics: accuracy (overall success rate), voluntary accuracy (success when the model voluntarily committed to an answer), overconfidence (incorrect committed answers), and underconfidence (continued questioning despite having enough information to answer correctly). 

Our 7B model improves accuracy and voluntary accuracy by 0.9 and 1.7 percentage points, respectively. The overconfidence rate drops from 36.1\% to 33.4\%, at a slight cost of 1.1 percentage points in underconfidence. Results are more mixed for the 14B model: accuracy remains essentially unchanged, voluntary accuracy decreases slightly from 64.6\% to 63.0\%, and overconfidence increases marginally by 1.5 percentage points.

Overall, ASIG transfers well to the medical domain at the 7B scale. The gains in voluntary accuracy and reduction in overconfidence align with our earlier finding that fine-tuning encourages more deliberate hypothesis-seeking, here reflected in the model's improved judgement about when to commit to an answer. The mixed 14B results are consistent with the weaker behavioural adaptation observed in the hypothesis-seeking analysis, suggesting that the larger model's stronger prior capabilities leave less room for fine-tuning to induce meaningful behavioural change.

\subsection{Capability Retention}
To ensure that fine-tuning does not degrade performance on other tasks or lead to failure modes such as catastrophic forgetting, we evaluate the baseline and fine-tuned models on three standard LLM benchmarks: ARC-Challenge \cite{clarkThinkYouHave2018}, HellaSwag \cite{zellersHellaSwagCanMachine2019}, and MMLU \cite{hendrycksAligningAIShared2021, hendrycksMeasuringMassiveMultitask2021}. 
As shown in \cref{tab:llm-benchmarks} in the Appendix, the fine-tuned models achieve near-identical scores to the base models across all three benchmarks, confirming that improved information-seeking abilities do not come at the cost of general language capabilities.

\section{Conclusion}
We introduced ASIG, a multi-turn GRPO-based fine-tuning method that amortises BED into LLM policies for sequential information gathering. By shifting information-seeking behaviour from inference-time optimisation into model weights, ASIG improves task performance while substantially reducing the computational overhead of sequential BED at deployment time.

Across two model scales, fine-tuning on 20 Questions consistently improves information-gathering performance on held-out categories, outperforming BED-LLM, a competitive inference-time BED baseline, in two of four settings while reducing inference cost by over $25\times$ overall. We further show partial transfer to MediQ, where ASIG improves clinical reasoning accuracy at the 7B scale without measurable degradation on standard language-understanding benchmarks. These results suggest that amortising information-seeking behaviour into model weights provides a practical and scalable approach to test-time information acquisition in LLMs.

Future work includes disentangling the effects of EIG and outcome-based rewards, evaluating broader sequential information-gathering benchmarks, and extending ASIG to reasoning models generating intermediate reasoning steps.

\section*{Acknowledgements}
We thank Tobias Bretschneider for his contributions to the development of the codebase and to the broader project. We are also grateful to Ari Weinstein for insightful discussions and valuable feedback throughout the course of this work. We thank Linus Filbry for making his BED-LLM implementation publicly available and for helpful discussions regarding its implementation details. Experiments reported in this work were supported by computational resources provided by the Ellison Institute of Technology, Oxford. The authors would also like to acknowledge the use of the University of Oxford Advanced Research Computing (ARC) facility in carrying out this work (DOI: 10.5281/zenodo.22558).

\bibliographystyle{icml2026_fogen}
\bibliography{references}

\newpage
\appendix
\section{Hyperparameters}
\begin{table}[H]
    \centering
    \caption{List of hyperparameters. All values are shared across the 7B and 14B model variants. Hyperparameters for BED-LLM and EIG calculation are taken from \citet{choudhuryBEDLLMIntelligentInformation2025}.}
    \begin{tabular}{lll}
\hline
Symbol                           & Name                                           & Value                                   \\ \hline
\multicolumn{3}{l}{\cellcolor[HTML]{EFEFEF}BED-LLM inference}                                                               \\
$N$                              & Belief set size                                & 15                                      \\
$M$                              & Candidate questions per turn                   & 15                                      \\
$\tau_f$                         & Belief filter threshold                        & 0.2                                     \\
$T_{\text{gen}}$                 & Generation temperature (beliefs \& candidates) & 1.3                                     \\
$T_a$                            & Answer temperature (likelihood)                & 0.7                                     \\
---                              & Max regeneration rounds                        & 3                                       \\
---                              & Monte Carlo samples for EIG                    & 15                                      \\
---                              & Retain consistent old beliefs                  & true                                    \\
\multicolumn{3}{l}{\cellcolor[HTML]{EFEFEF}Reward}                                                                          \\
$\alpha$                         & EIG reward weight curriculum                   & $0.0 \xrightarrow{\text{epoch 20}} 1.0$ \\
$\beta$                          & Outcome reward weight                          & 1.0                                     \\
$r_{\text{out}}(\text{Correct})$ & Correct-guess reward                           & +5                                      \\
$r_{\text{out}}(\text{Yes/No})$  & Step penalty                                   & -1                                      \\
$N$                              & Belief set size (reward model)                 & 15                                      \\
$\tau_f$                         & Belief filter threshold (reward model)         & 0.2                                     \\
$T_a$                            & Answer temperature (reward model)              & 0.7                                     \\
$E$                              & EIG rollout horizon                            & 1                                       \\
$O$                              & Outcome rollout horizon                        & 3                                       \\
$\gamma_{\text{eig}}$            & EIG discount factor                            & 1.0                                     \\
$\gamma_{\text{out}}$            & Outcome discount factor                        & 0.9                                     \\
\multicolumn{3}{l}{\cellcolor[HTML]{EFEFEF}GRPO training}                                                                   \\
---                              & Epochs                                         & 40                                      \\
---                              & Initial learning rate                          & $2 \times 10^{-6}$                      \\
$\beta_{\mathrm{KL}}$            & KL penalty coefficient                         & 0.0                                     \\
$G$                              & Group size (completions per prompt)            & 8                                       \\
---                              & Per-device batch size                          & 2                                       \\
---                              & Gradient accumulation steps                    & 8                                       \\
---                              & Max completion length (tokens)                 & 128                                     \\
---                              & Sampled entities per epoch                     & 100                                     \\
---                              & Dataset refresh (epochs)                       & 1                                       \\
---                              & Questions budget per game                      & 20                                      \\
\multicolumn{3}{l}{\cellcolor[HTML]{EFEFEF}LoRA}                                                                            \\
$r$                              & Rank                                           & 32                                      \\
$\alpha_{\text{LoRA}}$           & Scaling factor                                 & 64                                      \\
---                              & Dropout                                        & 0.05                                    \\
---                              & Target modules                                 & all attention + MLP projections         \\ \hline
\end{tabular}%

    \label{tab:hyperparameters}
\end{table}

\newpage
\section{Prompts}
\begin{prompt}{Proposer Agent System Prompt}{naive_agent_system}
You are playing a game of 20 Questions. Using all previous questions and answers:\\[6pt]
Generate the best question to help identify the target \{category\}. The question should be phrased so the answer is Yes or No. Print only the question - no numbering, punctuation, or extra text.
\end{prompt}

\begin{prompt}{Oracle Answerer System Prompt}{oracle_answerer_system}
You are playing the oracle answerer in a game of 20 Questions. The questioner has to guess a secret entity from the category: \{category\}. Your secret entity is: \{entity\}\\[6pt]
CRITICAL INSTRUCTIONS:\\
1. If the question directly identifies, names, or guesses ``\{entity\}" as the answer, respond ONLY with the word "Correct"\\
2. For yes/no questions about properties or characteristics, respond ONLY with "Yes" or "No"\\
3. Never add explanations, extra words, or punctuation\\[6pt]
Examples:\\
- "Is it \{entity\}?" \texttt{->} "Correct"\\
- "Is it a \{entity\}?" \texttt{->} "Correct"\\
- "Are you thinking of \{entity\}?" \texttt{->} "Correct"\\
- "I think it's \{entity\}" \texttt{->} "Correct"\\
- "Is the \{category\} a \{entity\}?" \texttt{->} "Correct"\\
- "Is it bigger than a car?" \texttt{->} "Yes" or "No" (depending on \{entity\})\\
- "Can you find it indoors?" \texttt{->} "Yes" or "No" (depending on \{entity\})\\
- "Is it man-made?" \texttt{->} "Yes" or "No" (depending on \{entity\})\\
- "Is it alive?" \texttt{->} "Yes" or "No" (depending on \{entity\})
\end{prompt}

\begin{prompt}{Likelihood Model System Prompt}{likelihood_model_system}
You are playing the answerer in a game of 20 Questions. Your chosen entity is: \{hypothesis\}\\
When asked a question, you must reply exactly ``Yes" or ``No", depending on if your chosen entity fulfills the question.
\end{prompt}

\begin{prompt}{Question Classifier System Prompt}{question_classifier_system}
You are classifying questions from a 20 Questions game.\\[6pt]
C - constraint-seeking: narrows the space of possible entities by asking about properties or categories (e.g. "Is it a mammal?", "Is it bigger than a car?")\\
H - hypothesis-seeking: asks whether the target IS a specific entity (e.g. "Is it a cat?", "Is it the Eiffel Tower?")\\
U - unknown: the question is ambiguous or does not fit either category\\[6pt]
Reply with ONLY the single letter: C, H, or U.
\end{prompt}

\begin{prompt}{Training Dataset Generation Prompt}{training_dataset_generation}
Role: You are a curator building a high-quality entity dataset for a 20 Questions game.

Task: Generate $\sim$50 entities for each of the categories listed below. Each entity must be guessable through $\sim$20 yes/no questions by an average adult player.

Categories \& quantities:

\begin{itemize}
    \item Musical instrument
    \item Fruit
    \item Organ
    \item Profession
    \item City
    \item Clothing
    \item Game
    \item Furniture
    \item Author
    \item Vegetable
    \item Sport
    \item Vehicle
    \item Electronic device
    \item TV show
    \item Celebrity
    \item Food chain
    \item University course
    \item Monument
    \item Cooking equipment
    \item River
\end{itemize}

Quality criteria - each entity must be:
\begin{enumerate}
    \item Recognizable - known to a general adult audience, not requiring specialist knowledge.
    \item Concrete \& unambiguous - a single, well-defined referent (e.g., "elephant," not "large mammal").
    \item Distinguishable - possesses a clear set of yes/no-answerable properties (size, habitat, function, era, material, etc.).
    \item Non-trivial - avoid items so generic they're guessed in 3 questions ("dog") or so obscure they're unguessable.
    \item Diverse within category - span subtypes, regions, eras, and difficulty levels; no near-duplicates.
\end{enumerate}

Output format: CSV

For reference, attached are example lists for the animals and plants category. Generate entities with a similar level of difficulty.
\end{prompt}

\newpage
\section{20 Questions Evaluation Datasets}
\begin{entitybox}{Animals dataset}{box:animals_dataset}
\begin{multicols}{4}
\begin{itemize}
  \item African elephant
  \item Bengal tiger
  \item Bald eagle
  \item Blue whale
  \item Red kangaroo
  \item Giant panda
  \item Snow leopard
  \item Green sea turtle
  \item American alligator
  \item Bottlenose dolphin
  \item Emperor penguin
  \item Great white shark
  \item Golden poison frog
  \item Honey bee
  \item Monarch butterfly
  \item Okapi
  \item Chimpanzee
  \item Arctic fox
  \item Komodo dragon
  \item Giraffe
  \item Cheetah
  \item Hammerhead shark
  \item Axolotl
  \item Orca
  \item Puffin
  \item Red panda
  \item Platypus
  \item Rhinoceros
  \item Tasmanian devil
  \item Wombat
  \item Sloth
  \item Blue-ringed octopus
  \item Manatee
  \item Narwhal
  \item Sea otter
  \item Coral snake
  \item King cobra
  \item Harpy eagle
  \item Lemur
  \item Koala
  \item Aye-aye
  \item Snowy owl
  \item Elk
  \item Wolverine
  \item Caracal
  \item Cassowary
  \item Quokka
  \item Pangolin
  \item Saiga antelope
  \item Galápagos tortoise
  \item Sumatran orangutan
  \item Red-eyed tree frog
  \item European badger
  \item Moose
  \item African grey parrot
  \item Scarlet macaw
  \item Black mamba
  \item Albatross
  \item Humpback whale
  \item Dugong
  \item Anaconda
  \item Kookaburra
  \item Coyote
  \item Brown bear
  \item Golden jackal
  \item Capybara
  \item Ibex
  \item Japanese macaque
  \item Kiwi
  \item Leafcutter ant
  \item Mantis shrimp
  \item Ocelot
  \item Peregrine falcon
  \item Quetzal
  \item Raccoon
  \item Sand cat
  \item Tarantula
  \item Uakari
  \item Vicuña
  \item Wildebeest
  \item Rock hyrax
  \item Yak
  \item Zebra
  \item Blue dragon nudibranch
  \item Chinchilla
  \item Dhole
  \item Electric eel
  \item Flying fox
  \item Gharial
  \item Horseshoe crab
  \item Indigo bunting
  \item Jerboa
  \item Kakapo
  \item Lionfish
  \item Markhor
  \item Nautilus
  \item Olive baboon
  \item Pika
  \item Quoll
  \item Rosy boa
\end{itemize}
\end{multicols}
\end{entitybox}
\captionof{mybox}{Animals dataset: 100 entities used in our evaluation. Dataset taken from \citet{choudhuryBEDLLMIntelligentInformation2025}.}
\label{box:animals_dataset}

\clearpage
\begin{entitybox}{Plants dataset}{box:plants_dataset}
\begin{multicols}{4}
\begin{itemize}
  \item Rose
  \item Oak
  \item Pine
  \item Palm
  \item Redwood
  \item Elm
  \item Cedar
  \item Juniper
  \item Hickory
  \item Willow
  \item Bamboo
  \item Hemlock
  \item Poplar
  \item Birch
  \item Cypress
  \item Magnolia
  \item Eucalyptus
  \item Maple
  \item Fir
  \item Beech
  \item Mahogany
  \item Spruce
  \item Dogwood
  \item Yew
  \item Larch
  \item Aspen
  \item Sycamore
  \item Hawthorn
  \item Holly
  \item Marigold
  \item Lily
  \item Primrose
  \item Hyacinth
  \item Lotus
  \item Crocus
  \item Anemone
  \item Carnation
  \item Iris
  \item Bluebell
  \item Lavender
  \item Orchid
  \item Poppy
  \item Pansy
  \item Dandelion
  \item Sunflower
  \item Narcissus
  \item Calendula
  \item Fuchsia
  \item Jasmine
  \item Chrysanthemum
  \item Lilac
  \item Ranunculus
  \item Azalea
  \item Gladiolus
  \item Hibiscus
  \item Daffodil
  \item Petunia
  \item Snowdrop
  \item Begonia
  \item Freesia
  \item Cornflower
  \item Hellebore
  \item Dahlia
  \item Peony
  \item Violet
  \item Camellia
  \item Hollyhock
  \item Alyssum
  \item Geranium
  \item Gardenia
  \item Scilla
  \item Primula
  \item Tuberose
  \item Poinsettia
  \item Amaryllis
  \item Impatiens
  \item Mimosa
  \item Forsythia
  \item Delphinium
  \item Bellflower
  \item Aster
  \item Trillium
  \item Hydrangea
  \item Zinnia
  \item Agave
  \item Aloe
  \item Cactus
  \item Saguaro
  \item Fern
  \item Philodendron
  \item Cyclamen
  \item Lupin
  \item Rudbeckia
  \item Yarrow
  \item Foxglove
  \item Wisteria
  \item Clematis
  \item Honeysuckle
  \item Allium
  \item Viburnum
\end{itemize}
\end{multicols}
\end{entitybox}
\captionof{mybox}{Plants dataset: 100 entities used in our evaluation. Dataset based on \citet{auzinaIntrinsicCreditAssignment2026}.}
\label{box:plants_dataset}

\newpage
\section{Extended Results}
\vspace{1.5em}
\begin{center}
\begin{minipage}{\textwidth}
    \centering
    \captionof{table}{\textbf{Success rate (mean $\pm$ std over episodes) for the baseline, ASIG, and BED-LLM on Qwen2.5-7B-Instruct and Qwen2.5-14B-Instruct across the animals and plants categories.} ASIG consistently improves over the baseline across all settings. BED-LLM outperforms ASIG on plants but falls below both ASIG and the baseline on the 14B animals category, suggesting it does not generalise robustly across model sizes and categories.}
    \label{tab:20_results}
    \begin{tabular}{@{}lllll@{}}
\toprule
Model                                 & Category & Baseline       & Ours (ASIG)             & BED-LLM                 \\ \midrule
\multirow{2}{*}{Qwen2.5-7B-Instruct}  & Animals  & 7.6 $\pm$ 1.8  & \textbf{21.0 $\pm$ 4.8} & 19.2 $\pm$ 4.0          \\
                                      & Plants   & 6.8 $\pm$ 1.6  & 16.6 $\pm$ 2.1          & \textbf{22.8 $\pm$ 2.2} \\
\multirow{2}{*}{Qwen2.5-14B-Instruct} & Animals  & 35.0 $\pm$ 2.0 & \textbf{39.2 $\pm$ 3.6} & 26.4 $\pm$ 4.8          \\
                                      & Plants   & 11.8 $\pm$ 1.9 & 13.6 $\pm$ 0.9          & \textbf{21.0 $\pm$ 4.1} \\ \bottomrule
\end{tabular}

\end{minipage}

\vspace{1cm}

\begin{minipage}{\textwidth}
    \centering
    \includegraphics[width=\textwidth]{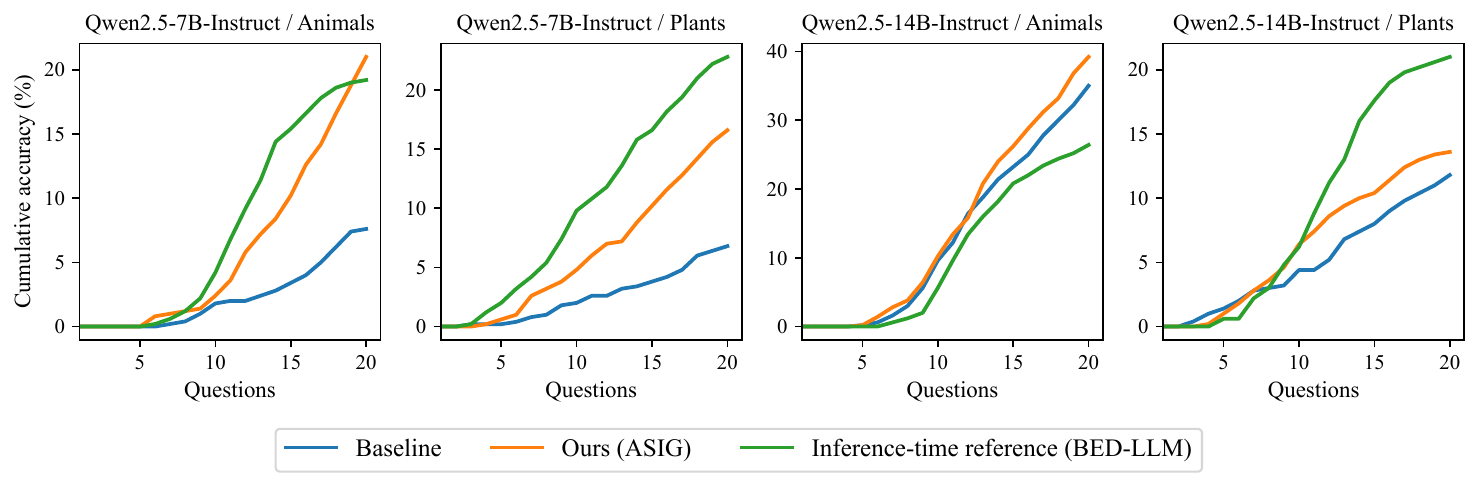}
    \captionof{figure}{\textbf{Cumulative fraction of games correctly identified within $N$ questions (average over 5 episodes) for Qwen2.5-7B-Instruct and Qwen2.5-14B-Instruct across the animals and plants categories.} ASIG outperforms the baseline at almost every turn and across the four settings. It also narrows the gap to or even outperforms the inference-time reference BED-LLM, despite using only a fraction of the inference-time compute.}
    \label{fig:20q_cumulative_accuracy}
\end{minipage}

\vspace{1cm}

\begin{minipage}{\textwidth}
    \centering
    \captionof{table}{\textbf{LLM benchmark scores for the baseline and ASIG on Qwen2.5-7B-Instruct and Qwen2.5-14B-Instruct across ARC-Challenge, HellaSwag, and MMLU.} ASIG causes no meaningful degradation across any benchmark or model size, confirming that general language capabilities are preserved.}
    \label{tab:llm-benchmarks}
    \begin{tabular}{@{}lllll@{}}
\toprule
Model                        & Variant    & ARC Challenge & HellaSwag & MMLU \\ \midrule
\multirow{2}{*}{Qwen2.5-7B}  & Baseline   & 43.6          & 65.4      & 68.7 \\
                             & Fine-tuned & 43.8          & 65.4      & 68.8 \\
\multirow{2}{*}{Qwen2.5-14B} & Baseline   & 50.0          & 75.6      & 76.8 \\
                             & Fine-tuned & 49.8          & 75.6      & 76.8 \\ \bottomrule
\end{tabular}%

\end{minipage}

\end{center}

\newpage
\section{Example Games}
\begin{tcolorbox}[
  colback=gray!5, colframe=black!60,
  title={Target: \texttt{Fir}},
  fonttitle=\bfseries,
]
\begin{minipage}[t]{0.48\linewidth}
  \textbf{Baseline} \hfill {\small\color{red!70!black}Failed (20 Q)}
  \vspace{2pt}\hrule\vspace{4pt}
  \begin{enumerate}[leftmargin=*, itemsep=1pt, label=\scriptsize\arabic*.]
    \item \qa{Does the plant require direct sunlight for at least 6 hours a day?}{\no}
    \item \qa{Does the plant have spines or thorns on its stems or leaves?}{\no}
    \item \qa{Is the plant native to a tropical or subtropical climate?}{\no}
    \item \qa{Does the plant shed its leaves annually?}{\no}
    \item \qa{Does the plant have flowers that are predominantly one solid color?}{\no}
    \item \qa{Does the plant have compound leaves with multiple leaflets?}{\no}
    \item \qa{Does the plant produce fruit that splits open when ripe?}{\no}
    \item \qa{Does the plant have a distinctive odor when crushed?}{\yes}
    \item \qa{Is the odor sweet and reminiscent of almond?}{\no}
    \item \qa{Does the plant have glossy leaves?}{\no}
    \item \qa{Does the plant have variegated leaves with different colors?}{\no}
    \item \qa{Does the plant produce small, inconspicuous flowers?}{\no}
    \item \qa{Does the plant have large, broad leaves?}{\no}
    \item \qa{Does the plant have needle-like leaves?}{\yes}
    \item \qa{Is the plant a conifer?}{\yes}
    \item \qa{Is the plant evergreen?}{\yes}
    \item \qa{Does the plant produce cones?}{\yes}
    \item \qa{Is the plant a pine tree?}{\no}
    \item \qa{Does the plant produce winged seeds?}{\yes}
    \item \qa{Is the plant a maple tree?}{\no}
  \end{enumerate}
\end{minipage}%
\hfill
\begin{minipage}[t]{0.48\linewidth}
  \textbf{Ours (ASIG)} \hfill {\small\color{green!50!black}Solved (16 Q)}
  \vspace{2pt}\hrule\vspace{4pt}
  \begin{enumerate}[leftmargin=*, itemsep=1pt, label=\scriptsize\arabic*.]
    \item \qa{Does the plant have leaves?}{\yes}
    \item \qa{Is the plant a deciduous tree?}{\no}
    \item \qa{Does the plant have flowers?}{\no}
    \item \qa{Is the plant a fern?}{\no}
    \item \qa{Does the plant have thorns?}{\no}
    \item \qa{Is the plant a grass or a grain?}{\no}
    \item \qa{Is the plant a shrub?}{\no}
    \item \qa{Is the plant a vine?}{\no}
    \item \qa{Is the plant a conifer?}{\yes}
    \item \qa{Is the plant a deciduous conifer?}{\no}
    \item \qa{Is the plant an evergreen conifer?}{\yes}
    \item \qa{Does the plant have needle-like leaves?}{\yes}
    \item \qa{Is the plant a pine?}{\no}
    \item \qa{Is the plant a cedar?}{\no}
    \item \qa{Is the plant a spruce?}{\no}
    \item \qa{Is the plant a fir?}{\yes}
  \end{enumerate}
\end{minipage}
\end{tcolorbox}
\captionof{mybox}{Example 20 Questions game with target \texttt{Fir}. The baseline fails to converge within the 20-question budget, while our fine-tuned 7B model identifies the target in 16 questions.}
\label{box:game_fir}

\clearpage
\begin{tcolorbox}[
  colback=gray!5, colframe=black!60,
  title={Target: \texttt{Honey bee}},
  fonttitle=\bfseries,
]
\begin{minipage}[t]{0.48\linewidth}
  \textbf{Baseline} \hfill {\small\color{red!70!black}Failed (20 Q)}
  \vspace{2pt}\hrule\vspace{4pt}
  \begin{enumerate}[leftmargin=*, itemsep=1pt, label=\scriptsize\arabic*.]
    \item \qa{Is the animal domesticated?}{\no}
    \item \qa{Does the animal live primarily on land?}{\yes}
    \item \qa{Is the animal a predator?}{\no}
    \item \qa{Does the animal live in herds or groups?}{\yes}
    \item \qa{Is the animal an herbivore?}{\yes}
    \item \qa{Is the animal larger than a car?}{\no}
    \item \qa{Is the animal a type of deer or antelope?}{\no}
    \item \qa{Does the animal have horns or antlers?}{\no}
    \item \qa{Is the animal a pig or related to pigs?}{\no}
    \item \qa{Is the animal a type of sheep or goat?}{\no}
    \item \qa{Is the animal a type of rodent?}{\no}
    \item \qa{Is the animal a horse or related to horses?}{\no}
    \item \qa{Is the animal a type of elephant or relative to elephants?}{\no}
    \item \qa{Is the animal a type of camel or related to camels?}{\no}
    \item \qa{Is the animal a type of primate?}{\no}
    \item \qa{Is the animal a type of wild boar or wild pig?}{\no}
    \item \qa{Is the animal a type of buffalo or bison?}{\no}
    \item \qa{Is the animal a type of llama or alpaca?}{\no}
    \item \qa{Is the animal a type of rhinoceros?}{\no}
    \item \qa{Is the animal a type of hippopotamus?}{\no}
  \end{enumerate}
\end{minipage}%
\hfill
\begin{minipage}[t]{0.48\linewidth}
  \textbf{Ours (ASIG)} \hfill {\small\color{green!50!black}Solved (11 Q)}
  \vspace{2pt}\hrule\vspace{4pt}
  \begin{enumerate}[leftmargin=*, itemsep=1pt, label=\scriptsize\arabic*.]
    \item \qa{Is the animal a mammal?}{\no}
    \item \qa{Is the animal cold blooded?}{\yes}
    \item \qa{Does the animal live primarily on land?}{\yes}
    \item \qa{Is the animal larger than a typical house cat?}{\no}
    \item \qa{Does the animal have wings?}{\yes}
    \item \qa{Can the animal fly?}{\yes}
    \item \qa{Is the animal a bird?}{\no}
    \item \qa{Is the animal a insect?}{\yes}
    \item \qa{Is the insect a butterfly or moth?}{\no}
    \item \qa{Is the insect a bee or wasp?}{\yes}
    \item \qa{Is the insect a honey bee?}{\yes}
  \end{enumerate}
\end{minipage}
\end{tcolorbox}
\captionof{mybox}{Example 20 Questions game with target \texttt{Honey bee}. The baseline fails to converge within the 20-question budget, while our fine-tuned 14B model identifies the target in 11 questions.}
\label{box:game_honey_bee}

\end{document}